\documentclass[10pt,twocolumn,letterpaper]{article}

\usepackage{cvpr}              

\definecolor{cvprblue}{rgb}{0.21,0.49,0.74}
\usepackage[pagebackref,breaklinks,colorlinks,allcolors=cvprblue]{hyperref}
\usepackage{multirow}
\usepackage{amssymb}
\usepackage{pifont}
\usepackage{colortbl}  
\usepackage{xcolor}
\newcommand{\cmark}{\ding{51}}%
\def\paperID{8878} 
\def\confName{CVPR}
\def\confYear{2025}

\newcounter{suppsection}

\title{SOLVE: Synergy of Language-Vision and End-to-End Networks for Autonomous Driving}

\author{Xuesong Chen\textsuperscript{1, 3, 4}\thanks{Equal contributions} \quad
Linjiang Huang\textsuperscript{2}\protect\footnotemark[1] \quad
Tao Ma\textsuperscript{1}  \quad
Rongyao Fang\textsuperscript{1} \quad
Shaoshuai Shi\textsuperscript{3} \thanks{Corresponding authors}  \quad
Hongsheng Li\textsuperscript{1, 4}\footnotemark[2] \\
\protect \textsuperscript{1}MMLab, CUHK \quad
\protect \textsuperscript{2} Institute of Artificial Intelligence, Beihang University \\
\protect \textsuperscript{3} Voyager Research, Didi Chuxing \quad 
\protect \textsuperscript{4} CPII under InnoHK \\
{\tt\small \{chenxuesong@link, hsli@ee\}.cuhk.edu.hk, ljhuang@buaa.edu.cn, shaoshuaics@gmail.com}
}

\begin{document}
\maketitle
\begin{abstract}
The integration of Vision-Language Models (VLMs) into autonomous driving systems has shown promise in addressing key challenges such as learning complexity, interpretability, and common-sense reasoning. However, existing approaches often struggle with efficient integration and real-time decision-making due to computational demands. In this paper, we introduce SOLVE, an innovative framework that synergizes VLMs with end-to-end (E2E) models to enhance autonomous vehicle planning. Our approach emphasizes knowledge sharing at the feature level through a shared visual encoder, enabling comprehensive interaction between VLM and E2E components. We propose a Trajectory Chain-of-Thought (T-CoT) paradigm, which progressively refines trajectory predictions, reducing uncertainty and improving accuracy. By employing a temporal decoupling strategy, SOLVE achieves efficient cooperation by aligning high-quality VLM outputs with E2E real-time performance. Evaluated on the nuScenes dataset, our method demonstrates significant improvements in trajectory prediction accuracy, paving the way for more robust and reliable autonomous driving systems.

\end{abstract}    
\section{Introduction}
\label{sec:intro}

Over the past two decades, autonomous driving has always been a key research focus, with the potential to significantly revolutionize transportation~\cite{muhammad2020deep,li2020deep,tian2024drivevlm}. Thanks to rapid advances in deep learning for perception~\cite{qi2017pointnet,lang2019pointpillars,sun2020scalability,li2022bevformer}, prediction~\cite{mozaffari2020deep,liu2021multimodal,gu2021densetnt}, and planning~\cite{song2020pip,ettinger2021large,caesar2021nuplan,teng2023motion,hu2023planning}, end-to-end autonomous driving~\cite{zhai2023rethinking,hu2023planning,jiang2023vad,chen2024vadv2,hwang2024emma,jiang2024senna} has gained prominence, where the entire driving process—from perception to planning—is learned within a unified framework. Typically, these methods predict future trajectories or control signals directly from sensor inputs, bypassing distinct decision-making steps. Through extensive data training, end-to-end approaches have demonstrated impressive planning capabilities, providing a streamlined and competitive alternative to traditional modular pipelines~\cite{chen2015hierarchical,kunz2015autonomous,noh2017decision}.


\begin{figure}[t]
\centering
\includegraphics[width=0.49\textwidth ]{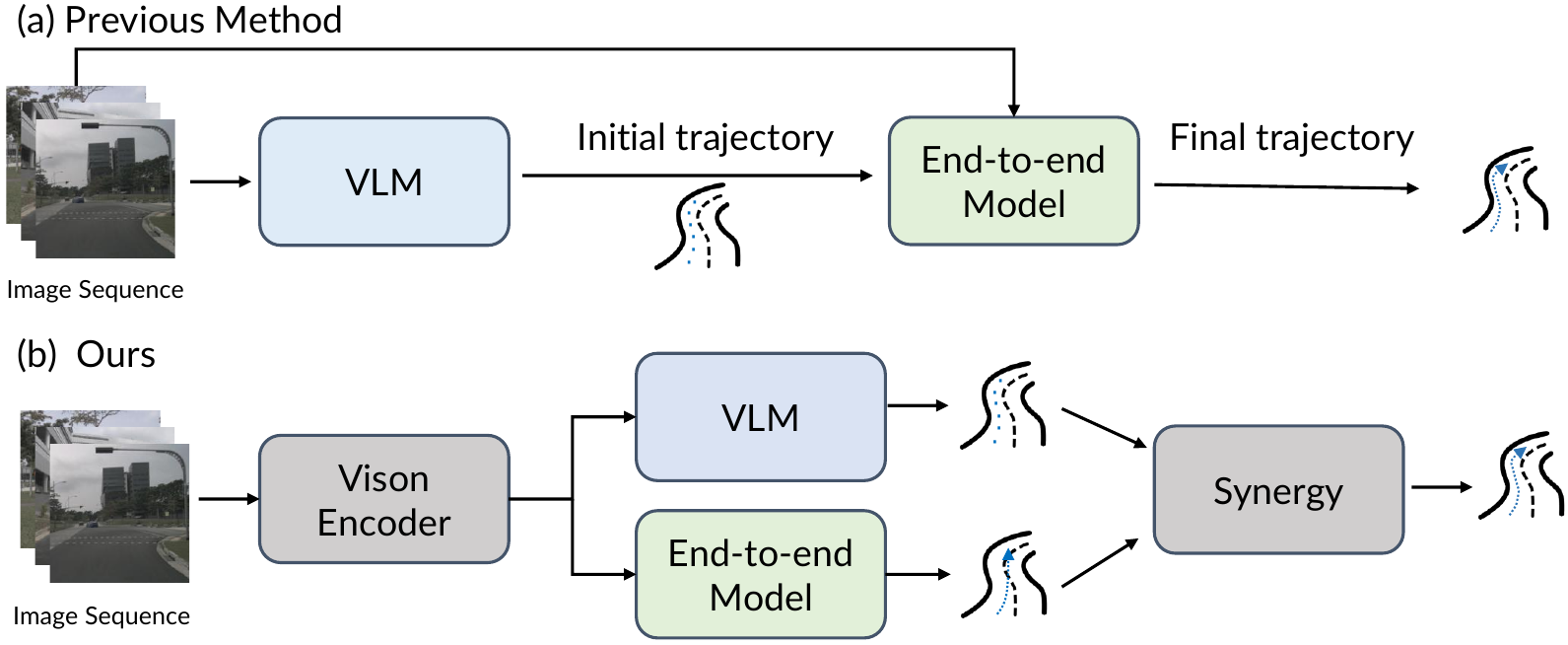}
\caption{Previous methods combine VLM and end-to-end networks through post-processing, while our method combines VLM and end-to-end networks through both feature-level synergy (shared visual encoder) and trajectory-level synergy.}
\vspace{-3mm}
\label{fig:teaser}
\end{figure}

End-to-end autonomous driving, while promising, still faces significant challenges, such as learning complexity~\cite{chen2024end}, insufficient common-sense reasoning~\cite{mao2023language,fu2024drive}, limited interpretability~\cite{jing2022inaction,shao2023safety}, and causal confusion~\cite{de2019causal,li2024exploring}. For instance, current approaches may misinterpret trucks carrying traffic cones as roadblocks or mistake billboard images of vehicles for real obstacles, resulting in unnecessary braking.
Recent advancements in large Vision-Language Models (VLMs)~\cite{alayrac2022flamingo,li2023blip,liu2024visual} offer a potential solution to these challenges. VLMs, through extensive training on diverse datasets, can simplify learning complexity by providing a more intuitive understanding of various scenarios. Additionally, the logical and commonsense knowledge provided by VLMs enhances autonomous driving systems, allowing them to better interpret driving environments and make safer decisions in complex situations. However, VLMs still face limitations in areas such as efficiency and 3D perception, which pose challenges for their integration into an end-to-end autonomous driving framework.

Some initial explorations have been made to leverage VLMs within the realm of autonomous driving. For example, DriveVLM~\cite{tian2024drivevlm} introduces a dual-system architecture that integrates the classic perception-prediction-planning (3P) pipeline with VLMs. As shown in Figure~\ref{fig:teaser}(a), the VLM and the end-to-end (E2E) model operate independently and cooperate at the trajectory level, \eg, using the E2E planner to refine the trajectory predictions made by the VLM. However, this strategy limits information and knowledge sharing between the VLM and the E2E model. For example, the E2E model cannot leverage the high-level reasoning abilities of VLM to pay more attention to critical objects or understand their intents, which is crucial for collision avoidance. Similarly, the VLM cannot integrate the 3D perception capabilities of the E2E model for precise scene comprehension, which is essential for accurate planning.

In this work, we introduce SOLVE, an innovative framework that promotes synergy between the VLM and the E2E model, emphasizing both knowledge and planning integration.
As Figure~\ref{fig:teaser} shows, 
besides the trajectory-level cooperation, which is already explored in previous approaches~\cite{tian2024drivevlm},
our method enhances feature-level knowledge sharing through a shared vision encoder, facilitating comprehensive collaboration between the VLM and the E2E model.
Leveraging the Q-Former design, the shared encoder efficiently compresses surrounding foreground objects (e.g., cars, pedestrians), background maps (e.g., lane centerlines), and environmental cues (e.g., traffic lights, weather, time of day) into unified visual tokens, reducing redundant feature extraction. Consequently, the Q-Former is designed to operate feature collection and alignment with various perception queries sequentially. This structural alignment enhances consistency across vision perception and control domains, significantly facilitating knowledge sharing and easing the difficulties of scene understanding.


To boost precision and efficiency in trajectory prediction, we meticulously optimize the VLM and the E2E model. To address the challenges of VLM in directly generating fine-grained trajectories in an auto-regressive manner, we introduce the Trajectory Chain-of-Thought (T-CoT). This new paradigm progressively refines trajectories by using a predefined trajectory bank and chain-style reasoning. Initially, a trajectory adapter transforms each candidate trajectory from the trajectory bank into trajectory tokens that align with text and visual tokens, enabling the network to select the most suitable trajectory. Subsequently, the selected trajectory serves as a foundation for the next round of refinement, ultimately generating the final trajectory.

Given that VLM can produce high-quality trajectories but at a significant computational cost, we propose a temporal decoupling strategy to enable cooperation between the VLM and the E2E model. Unlike previous methods \cite{chen2025asynchronous}, which perform feature-level asynchronization, our approach operates at the trajectory level, specifically aiming to enhance trajectory accuracy. The framework implements a memory-based mechanism where VLM-generated trajectories are stored and subsequently accessed by the E2E network as initialization priors, accommodating a temporal offset from the current state. This approach leverages priors of VLM to asynchronously enhance E2E network performance while maintaining real-time system operation capabilities.

In summary, our contributions are three-fold:
\begin{itemize}
    \item We propose SOLVE, an innovative framework that promotes synergy between the VLM and the E2E model, emphasizing both knowledge and trajectory integration.
    \item We propose the Trajectory Chain-of-Thought (T-CoT),  which progressively refines trajectories through a predefined trajectory bank and chain-style reasoning. 
    \item  We demonstrate the effectiveness of our proposed approach through extensive experiments, and our framework achieves state-of-the-art results for open-loop planning on the nuScenes benchmark.
\end{itemize}

\section{Related Work}
\label{sec:related_work}

\subsection{End-to-end Autonomous Driving}

In the early stage, autonomous driving systems often follow a modular design and are equipped with rule-based planners. This design hurdles the system's generalization ability and leads to limited performance. In contrast, 
end-to-end driving is a fully differentiable system from sensors to control signals, optimizing the entire system to directly map perception inputs to planning outputs. Some early works primarily rely on data-driven approaches to train neural networks for end-to-end autonomous driving. These methods show promise, despite challenges with interpretability. The growth in driving data further promotes end-to-end autonomous driving,
with models like UniAD~\cite{hu2023planning} and PARA-Drive~\cite{weng2024drive} using Transformers for multi-task learning like object tracking and planning, enhancing performance through additional supervision. Furthermore, VAD~\cite{jiang2023vad} introduces vectorized scene representation for better accuracy and speed, with VAD-\emph{v}2~\cite{chen2024vadv2} further improving planning by capturing uncertainty.
Besides, methods such as AD-MLP~\cite{zhai2023rethinking} employ a simple multi-layer perceptron (MLP) that relies solely on ego vehicle status information, highlighting the potential over-reliance on ego status in simpler driving scenarios while also revealing limitations in utilizing perception data effectively.
BEV-Planner~\cite{li2024ego} integrates various perception inputs in a bird's-eye view (BEV) format, aiming to enhance planning capabilities in more complex traffic situations by leveraging richer environmental context.

\subsection{VLMs for Autonomous Driving}

Nevertheless, end-to-end autonomous driving often suffers from limitations, such as  learning complexity~\cite{chen2024end}, insufficient common-sense reasoning~\cite{mao2023language,fu2024drive}, limited interpretability~\cite{jing2022inaction,shao2023safety}, and causal confusion~\cite{de2019causal,li2024exploring}.
There is an urgent need to improve these systems for real-world applications. Integrating Vision-Language Models (VLMs) into autonomous driving systems has gained significant attention due to their potential to enhance explainability, reasoning, and generalization in driving tasks. For instance, models like DriveGPT4~\cite{xu2024drivegpt4} utilize VLMs to predict control signals while simultaneously providing explanations for vehicle actions through an iterative question-and-answer format. Similarly, Drive Anywhere~\cite{wang2024drive} employs patch-aligned feature extraction from multi-modal large language models (MLLMs) to facilitate text-based driving decision-making, while OmniDrive~\cite{wang2024omnidrive} introduces a 3D vision-language model architecture aimed at improving reasoning and planning capabilities. Other innovative approaches, such as DriveLM~\cite{sima2023drivelm}, leverage graph-based visual question-answering (VQA) techniques. 
In a complementary manner, Drive-with-LLMs~\cite{chen2024driving} encodes perception data into latent space using a Transformer network, which is then processed by an LLM to forecast future trajectories.
However, most of these VLM-based approaches requires a large computation cost, which cannot achieve real-time decision making. 
DriveVLM~\cite{tian2024drivevlm} stands out as the first model to merge VLMs with end-to-end architectures, predicting low-frequency trajectories through chain-of-thought reasoning that are refined by an end-to-end model to produce final planning outputs. Despite its promising performance, it only fosters cooperation at the trajectory level, overlooking the high-level complementarity between the VLM and the end-to-end model.

\section{SOLVE}
\begin{figure*}[t!]
\centering
\includegraphics[width=0.99\textwidth ]{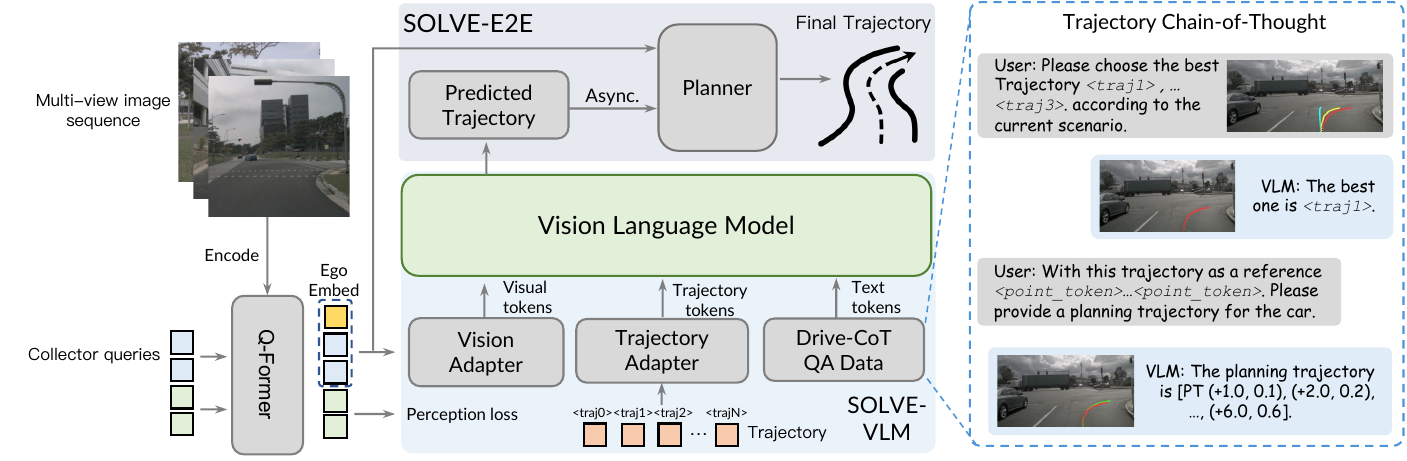}
\caption{The overall framework of the proposed SOLVE.}
\vspace{-3mm}
\label{fig:pipeline}
\end{figure*}
In this section, we provide a detailed introduction to SOLVE. Figure~\ref{fig:pipeline} illustrates the overall architecture. We utilize a Sequential Q-Former (SQ-Former) to obtain rich and condensed visual cues from sequences image input (Section~\ref{sec:perception}). These visual cues are shared between the Vision-Language Model (VLM) and the end-to-end (E2E) model, enabling efficient feature-level knowledge integration.
Additionally, we introduce a Trajectory Chain-of-Thought (T-CoT) mechanism (Section~\ref{sec:t-cot}) that refines trajectory predictions from coarse to fine through chain-style reasoning. By aligning the trajectory predictions of the VLM with E2E real-time processing in a time-decoupled cooperation manner (Section~\ref{sec:asynchronous}), SOLVE achieves the balance between accuracy and computational efficiency. Finally, we introduce our multi-stage training strategy (Section~\ref{sec:training_strategy}) to further promote feature-level synergy and trajectory prediction.
\begin{figure}[t]
\centering
\includegraphics[height=4.5cm,width=0.47\textwidth ]{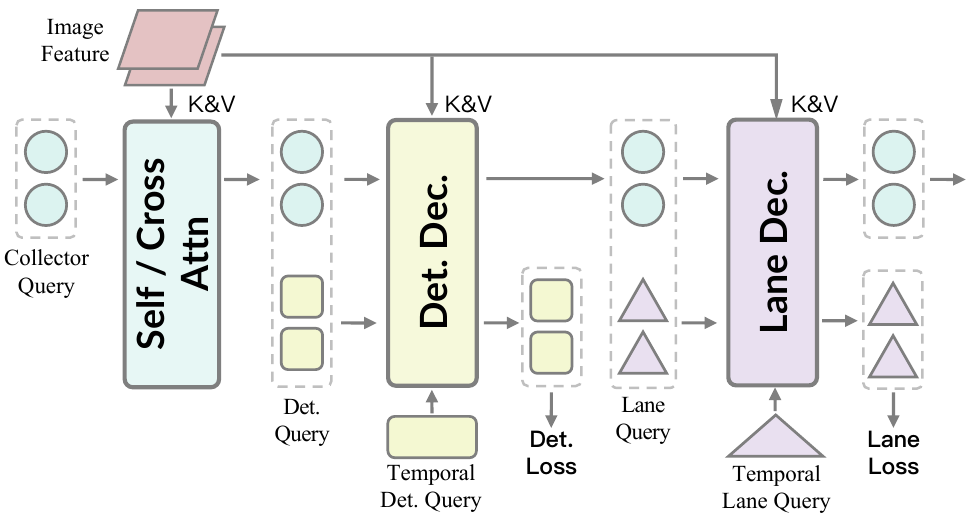}
\caption{The detail of the proposed SQ-Former. We first capture the static cues from multi-view images and then sequentially align the model with different perception tasks.}
\vspace{-3mm}
\label{fig:q-former}
\end{figure}

\subsection{SQ-Former} \label{sec:perception}

Visual cues play a pivotal role in autonomous driving, offering foreground information, scene map information, and environmental cues. These elements help the system to accurately predict vehicle behavior and plan safe routes, enhancing overall decision-making and navigation capabilities. However, abundant visual information can also introduce interference in autonomous driving, such as irrelevant background clutter or reflections. Effectively filtering and extracting useful visual information ensures accuracy and efficiency for autonomous driving.

To harness the relevant information from observations, we categorize useful visual cues into three groups, as shown in Figure~\ref{fig:q-former}. The first group includes scene-level static or holistic environmental details, such as weather, time of day, and traffic conditions. The second group concentrates on dynamic participants on the road, such as vehicles and pedestrians. The third group contains dynamic map cues, \eg, lanes. Here, we present Sequential Q-Former (SQ-Former), a Q-Former-style architecture that sequentially compresses these groups of visual cues into a fixed number of condensed visual queries, aligning perception with reasoning and planning.
Specifically, the SQ-Former is built upon an image encoder, mapping the multi-view images into high-level features $\mathbf{F}_i \in \mathbb{R}^{N \times C \times H \times W}$, where $N, C, H, W$ denotes the number of views, channels, height and width, respectively. The extracted features are forwarded into the SQ-Former to progressively collect both static and dynamic visual cues from observations. Inspired by StreamPETR~\cite{wang2023exploring}, we begin with a set of collector queries $\mathbf{Q}_c$. These queries are input into a transformer decoder to capture static cues of the first group, which can be formulated as:
\begin{equation}
    \mathbf{Q}_c^{(s)} = f_s(\mathbf{Q}_c, \mathbf{F}_i + \mathbf{P}_i),
\end{equation}
where $\mathbf{P}_i$ denotes the 3D position embedding. $f_s$ stands for the transformer decoder, which includes interleaved self-attention and cross-attention layers to promote interaction between queries and extract informative static cues, respectively. Afterward, the updated collector queries $\mathbf{Q}_c^{(s)}$ are concatenated with task-specific queries $\mathbf{Q}_{\blacktriangle}$, where $\blacktriangle$ represents different tasks, such as object detection (d) or map understanding (m). These queries are then fed into separate transformer decoders.
In SQ-Former, we sequentially conduct this process, following a pre-defined order of perception tasks, which can be written as:
\begin{align}
&\mathbf{Q}_c^{(s,\blacktriangle[1:n])},
\tilde{\mathbf{Q}}_{\blacktriangle[n]} \notag \\ &= 
f_{\blacktriangle[n]}\big([\mathbf{Q}_c^{(s,\blacktriangle[1:n-1])}, 
\mathbf{Q}_{\blacktriangle[n]}], \mathbf{F}_i + \mathbf{P}_i, \mathcal{M}\big),
\end{align}
where $[\cdot]$ denotes the concatenation operation, and $f_{\blacktriangle[n]}$ is the transformer decoder specific to the \emph{n}-th task. We introduce the memory bank $\mathcal{M}$, following StreamPETR~\cite{wang2023exploring}, to incorporate the temporal dynamics of the driving scenario. In this framework, $\mathbf{Q}c^{(s,\blacktriangle[1:n])}$ and $\tilde{\mathbf{Q}}{\blacktriangle[n]}$ represent the collector queries enriched with visual cues for tasks $1$ through $n$, and the updated task-specific queries, respectively. These updated queries are then used to align with the \emph{n}-th task. The collector queries are sent to a vision adapter to match the dimensions of the VLM, with an ego token appended for trajectory prediction.

According to our experiments, the sequential design ensures that SQ-Former captures rich, condensed visual cues from various perception tasks, enhancing synergy between them. For example, compared to OmniDrive~\cite{wang2024omnidrive} that uses 512 queries, our method achieves superior performance with only 384 queries, significantly reducing the computational cost of the following VLM. 

\subsection{Trajectory Chain-of-Thought} \label{sec:t-cot}

AD-MLP~\cite{zhai2023rethinking} has demonstrated that even a simple planner can achieve moderate results using only ego status. However, it lacks the ability to reason and integrate logical knowledge, which is crucial for a safe and reliable autonomous driving system. To address this limitation, we introduce VLM to enhance the system's reasoning abilities.


Specifically, we provide the VLM with visual and ego inputs, along with high-quality question-answering (QA) pairs that encompass perception, reasoning, and planning~\cite{wang2024omnidrive}.
Unlike previous methods, we meticulously develop a Trajectory Chain-of-Thought (T-CoT) process for the planning phase, employing a coarse-to-fine procedure for trajectory prediction. Notably, there are several methods, such as DriveCoT~\cite{wang2024drivecot}, DriveLM~\cite{sima2023drivelm} and DriveVLM~\cite{tian2024drivevlm}, have demonstrated the effectiveness of CoT in autonomous driving. However, their reasoning chain spans from perception to planning, requiring multiple rounds of reasoning and lengthy textual output, resulting in significant computational costs. In contrast, our trajectory CoT operates at the trajectory level, refining trajectories in several steps and requiring minimal output.

To make the T-CoT work at the trajectory level, we first construct a trajectory bank containing 36 candidate trajectories (with corresponding ego status and history trajectory) by clustering training set trajectories using the \emph{k}-means algorithm for each navigation command (forward, left-turn and right-turn). Next, we utilize the current ego status and historical trajectory as a feature vector, calculate its similarity with the corresponding feature vectors to the clustered trajectories, and then use a top-k operation to select the $k_l$ trajectories that are closest to the current feature vector.
These are combined with a trajectory predicted by an MLP, which uses only ego information and historical data as input~\cite{zhai2023rethinking}, resulting in $k_l+1$ candidate trajectories. We introduce clustered trajectories because, although the coarse trajectories are already acceptable, some predicted trajectories still deviate significantly from the ground truth. Retrieving clustered trajectories provides a pull-back mechanism to correct these inaccuracies. With the trajectory bank, the VLM is tasked with progressively outputting predictions in two stages: trajectory selection and trajectory refinement.

\begin{figure}[t]
\centering
\includegraphics[width=0.47\textwidth ]{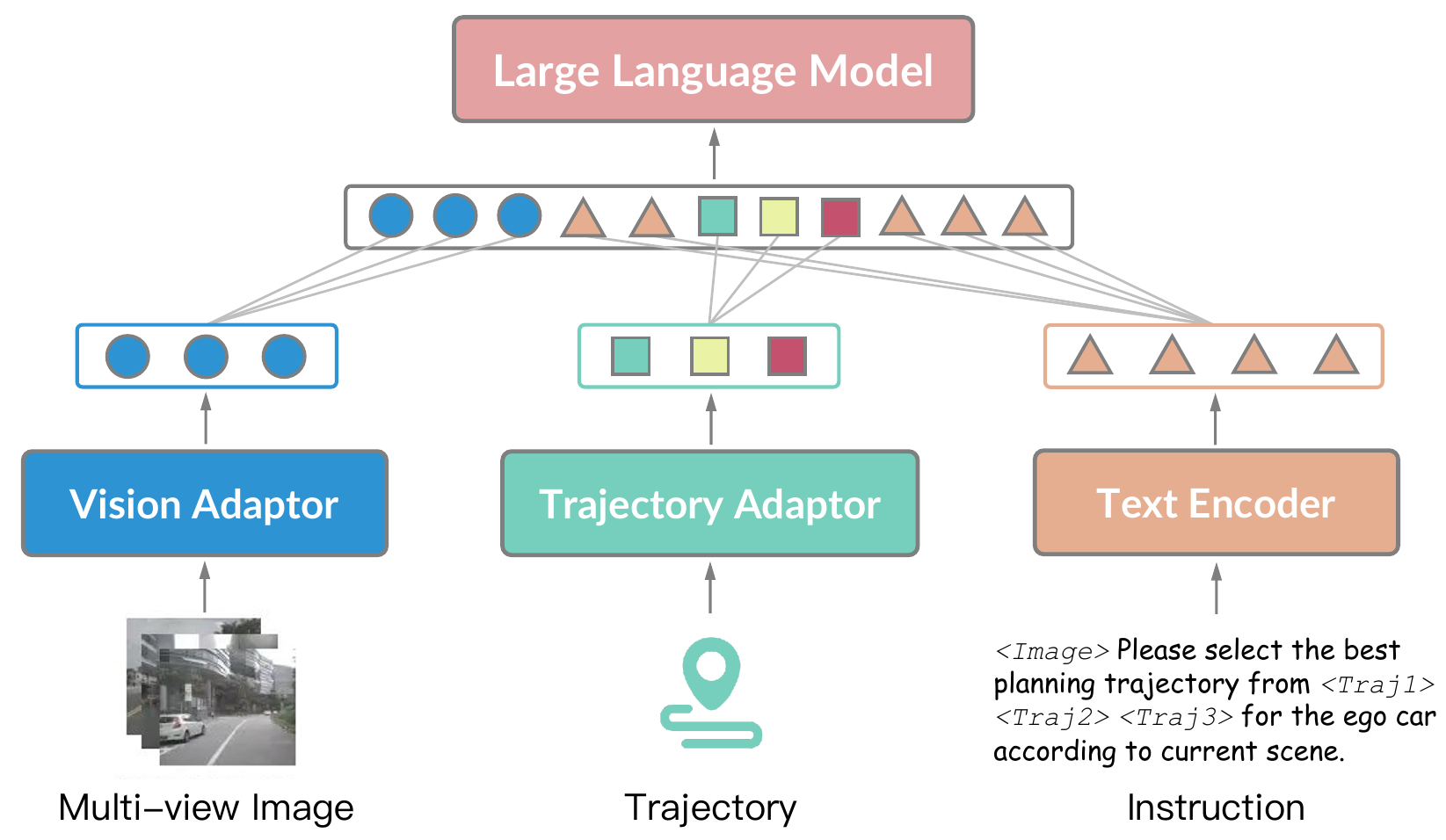}
\caption{The illustration of combination of the proposed trajectory tokens with image tokens and text tokens for the large language model-based planning. }
\label{fig:traj_token}
\end{figure}

\textbf{Trajectory selection} prompts the VLM to choose the most accurate trajectory from the candidates, based on ego and environmental information. To reduce computational costs, the endpoint of each candidate trajectory is encoded into a single token through the proposed trajectory adapter and inserted into the textual prompt as follows: \emph{Here are predefined trajectories \textless Traj$_1$\textgreater, \textless Traj$_2$\textgreater, $\cdots$, \textless Traj$_{k_l+1}$\textgreater. 
Please select the best trajectory for the ego car according to the current scenario.}, as shown in Figure~\ref{fig:traj_token}.


\textbf{Trajectory refinement} involves generating a refined trajectory after selecting the reference trajectory. The reference trajectory consists of waypoints, denoted by \( W = \{ w_1, w_2, \ldots, w_n \} \), where \( w_i = (x_i, y_i) \), representing the vehicle’s path over a future period with predetermined intervals. Each waypoint of the reference trajectory is encoded into a waypoint token and injected into the textual prompt as follows: \emph{With the selected trajectory as a reference \textless Point$_1$\textgreater, \textless Point$_2$\textgreater, $\cdots$, \textless Point$_n$\textgreater, please provide the planning trajectory for the ego car, which has a velocity of SPEED m/s and an acceleration of ACCEL $\text{m}/{\text{s}}^{2}$.}


\subsection{Time-decoupled Synergy Prediction} \label{sec:asynchronous}

Our planning-oriented synergy system operates at both feature-level (shared SQ-Former feature) and the trajectory-level, aiming to further enhance final trajectory accuracy. This strategy combines the advanced capabilities of VLMs in generating high-accuracy trajectories with the efficiency of a streamlined E2E model.

Specifically, we deploy a simple planner that incorporates cross-attention mechanisms between multiple planning queries and collector queries and then it employs an MLP to project these queries into the final output trajectories. The initialization of the planning queries are performed by encoding position embedding from clustering trajectories from the training set, following SparseDrive~\cite{sun2024sparsedrive}. The incorporation between the SQ-Former and the simple planner constructs an E2E model, a pipeline frequently adopted in prior systems.

Upon generating a trajectory from the VLM, we put it into memory. The end-to-end network can obtain the latest VLM predicted trajectory from the memory (with a delay from the current time) as an additional trajectory initialization. Specifically, we task the VLM with predicting trajectories over extended time horizons (longer than both the VLM's latency and the prediction horizon required by the end-to-end model). Consequently, despite the VLM's latency, the E2E model can still asynchronously leverage the VLM's long-term trajectories to obtain additional high-quality trajectory initialization and get the final collaborative predicted trajectory.


\subsection{Training Strategy} \label{sec:training_strategy}


We employ a multi-stage training strategy for SOLVE. In the first stage, we employ the same QA pairs as proposed by OmniDrive~\cite{wang2024omnidrive}, prompting the VLM by incorporating LoRA layers~\cite{hu2021lora} to address inquiries related to scene description, perception, prediction, and planning. Through QA training, the collector queries in the visual encoder are optimized to comprehensively encode vital scene information while maintaining efficiency.


Subsequently, we train a trajectory adapter to ensure that it can encode predefined trajectories into a feature space compatible with the VLM. We then train the end-to-end planning head while freezing the previously trained SQ-Former, vision adapter, and trajectory adapter.

In the subsequent phase, leveraging these pre-trained weights, we jointly optimize the VLM and end-to-end planning head without freezing any components. During this phase, we employ the proposed T-CoT to augment the reasoning capabilities of the VLM.

Finally, our trajectory-level synergy strategy involves exclusively refining the end-to-end planner by leveraging the trajectory forecasted by the VLM in historical frames as an additional initialized planning query for the current frame. 

\section{Experiments}

\subsection{Dataset and Implementation Details}

\begin{table*}[!t]
\begin{center}
\scalebox{0.9}{
\setlength{\tabcolsep}{6pt}
\begin{tabular}{l|c|cccc|cccc}
\toprule
\multirow{2}{*}{Method} &
\multirow{2}{*}{Ego Status} &
\multicolumn{4}{c|}{L2 (m) $\downarrow$} & 
\multicolumn{4}{c}{Collision (\%) $\downarrow$} \\
 & & 1s & 2s & 3s &\cellcolor{gray!30}Avg. & 1s & 2s & 3s& \cellcolor{gray!30}Avg.\\
\midrule
UniAD~\cite{hu2023planning} &-& 0.59 & 1.01 & 1.48 & \cellcolor{gray!30}1.03& 0.16 & 0.51 & 1.64 & \cellcolor{gray!30}0.77  \\
UniAD~\cite{hu2023planning} &\cmark& 0.20 & 0.42 & 0.75& \cellcolor{gray!30}0.46 & 0.02 & 0.25 & 0.84& \cellcolor{gray!30}0.37  \\
\midrule
VAD-Base~\cite{jiang2023vad} &-& 0.69 & 1.22 & 1.83 & \cellcolor{gray!30}1.25 & 0.06 & 0.68 & 2.52 & \cellcolor{gray!30}1.09  \\
VAD-Base~\cite{jiang2023vad} &\cmark& 0.17 & 0.34 & 0.60 & \cellcolor{gray!30}0.37 & 0.04 & 0.27 & 0.67 & \cellcolor{gray!30}0.33  \\
\midrule
AD-MLP~\cite{zhai2023rethinking} & - & 0.15 & 0.32 & 0.59  & \cellcolor{gray!30}0.35&\textbf{0.00} & {0.27} & 0.85& \cellcolor{gray!30}0.37 \\
\midrule
BEV-Planner~\cite{li2024ego} &- & 0.30 & 0.52&0.83 & \cellcolor{gray!30}0.55 & 0.10 & 0.37 & 1.30 & \cellcolor{gray!30}0.59\\
BEV-Planner++~\cite{li2024ego} &\cmark & 0.16 & 0.32& 0.57 & \cellcolor{gray!30}0.35& \textbf{0.00} & 0.29 & 0.73 & \cellcolor{gray!30}0.34 \\
\midrule
DriveVLM~\cite{tian2024drivevlm} &\cmark & 0.18 & 0.34 & 0.68 & \cellcolor{gray!30}0.40 & - & - & - & \cellcolor{gray!30}- \\
OmniDrive~\cite{wang2024omnidrive} &\cmark & 0.14 & 0.29 & 0.55 & \cellcolor{gray!30}0.33 & \textbf{0.00} & \textbf{0.13} & 0.78 & \cellcolor{gray!30}0.30  \\
EMMA~\cite{hwang2024emma} &\cmark & 0.14 & 0.29 & 0.54 & \cellcolor{gray!30}0.32 & - & - & - & \cellcolor{gray!30}- \\
DriveVLM-Dual~\cite{tian2024drivevlm} &\cmark & 0.15 & 0.29 & 0.48 & \cellcolor{gray!30}0.31 & - & - & - & \cellcolor{gray!30}- \\
Ours-E2E &\cmark & 0.14 & 0.28 & 0.50 & \cellcolor{gray!30}0.31 & 0.04 & 0.17 & 0.68 & \cellcolor{gray!30}0.30 \\
Ours-E2E (Async) &\cmark & 0.13 & 0.27 & 0.50 & \cellcolor{gray!30}0.30 & 0.04 & 0.17 & 0.65 & \cellcolor{gray!30}0.29 \\
Ours-VLM &\cmark & \textbf{0.13} & \textbf{0.25} & \textbf{0.47} & \cellcolor{gray!30}\textbf{0.28} & \textbf{0.00} & 0.16 & \textbf{0.43} & \cellcolor{gray!30}\textbf{0.20} \\

\bottomrule
\end{tabular}}
\vspace{-4mm}
\end{center}
\caption{Performance comparison of various methods on the NuScenes dataset for open-loop planning. Methods are evaluated with or without ego status consideration. The proposed methods, particularly SOLVE-VLM, demonstrate superior accuracy and safety, achieving the lowest average L2 error and collision rates.}
\label{tab:sota-plan}
\vspace{-3mm}
 \end{table*}

\noindent
\textbf{NuScenes Dataset.} We utilize the nuScenes~\cite{caesar2020nuscenes} dataset to train and evaluate our method. This comprehensive, large-scale driving dataset is tailored for urban environments, featuring 1,000 driving scenes, each approximately 20 seconds long. It encompasses diverse sensor data, including surround-view camera images and LiDAR point clouds, along with extensive annotations such as 3D bounding boxes, ego vehicle trajectories, and road maps. 
For training of VLM, we employ the OmniDrive-nuScenes~\cite{wang2024omnidrive} dataset, a planning-oriented extension of the nuScenes dataset with additional annotations, including perception, prediction, and planning-related QA. Specifically, the dataset features two distinct QA paradigms: Offline QA, which focuses on scene-level reasoning through tasks such as scene description, attention analysis, and decision validation using language models; and Online QA, which emphasizes 3D spatial understanding through tasks like 2D-to-3D grounding, proximity analysis, and lane-based object detection using geometric labels.

\noindent
\textbf{Evaluation Metrics.} Following previous works~\cite{li2024ego}, we employ displacement error (L2 distance) and collision rate as the primary metrics.

\noindent
\textbf{Implementation Details.} We use LLaVA v1.5~\cite{liu2024visual}, whose component of language model is based on LLaMA-7B, as our default large vision-language model, demonstrating exceptional performance in tasks such as question answering, visual localization, and text recognition. We replace its vision encoder with our SQ-Former. For the SQ-Former, we employ a predefined task order of image $\rightarrow$ 3D detection $\rightarrow$ lane detection. The image encoder is initialized using EVA-02-L~\cite{fang2024eva}. For the detection decoder and lane decoder, we adopt the architecture from StreamPETR~\cite{wang2023exploring}. The numbers of object queries, lane queries, and the sequential collector queries are set to 900, 300, and 384, respectively. 

\begin{figure*}[t]
\centering
\includegraphics[width=0.98\textwidth ]{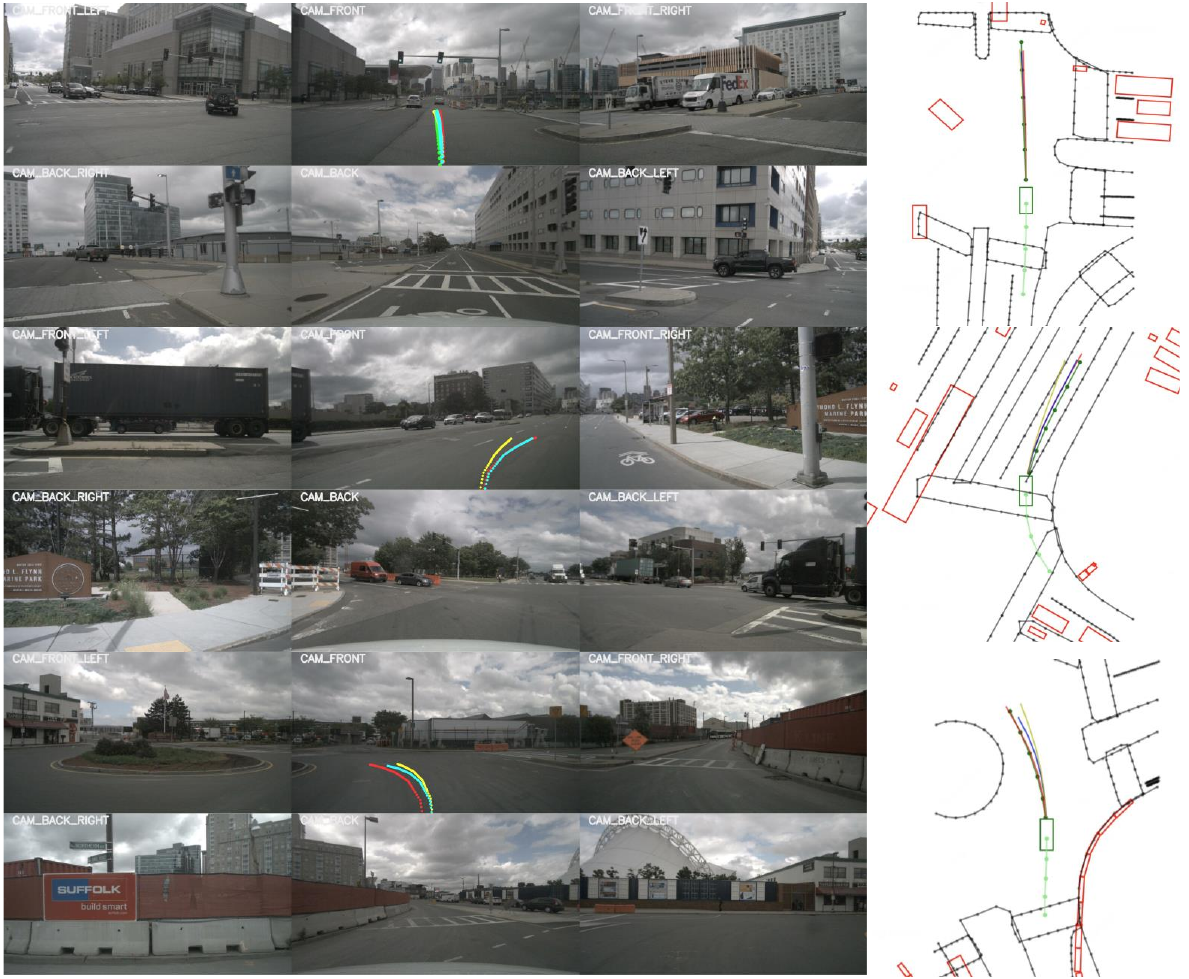}
\vspace{-3mm}
\caption{Qualitative results of SOLVE, where red lines, blue lines and yellow lines mean the planning results from VLM, E2E-Async and E2E modules. The ego car (green box) and ground truth trajectory (green line) are shown in the right BEV images.}
\vspace{-5mm}
\label{fig:vlm_input}
\end{figure*}

\subsection{Comparison with State-of-the-art Methods}
As shown in Table~\ref{tab:sota-plan}, SOLVE-E2E and SOLVE-VLM achieved SOTA results in their respective tracks. Specifically, among end-to-end networks, SOLVE-E2E surpassed UniAD~\cite{hu2023planning}, VAD~\cite{jiang2023vad}, and BEV-planner~\cite{li2024ego} by 0.15, 0.06, and 0.04, respectively, in terms of L2 (m) error, which shows that SOLVE can better imitate the driving behavior of human experts. On the other hand, SOLVE achieves 0.07\%, 0.03\%, and 0.07\% lower collision rates than UniAD, VAD, and BEV-planner, respectively, which shows that it has more accurate spatial judgment ability than others and can avoid other objects as much as possible when planning trajectories. It is worth noting that SOLVE-E2E achieved similar performance to DriveVLM-dual. This proves the superiority of the joint training of SOLVE's shared visual encoder. Specifically, other end-to-end methods only train visual encoders through detection or segmentation tasks, but lack scene understanding capabilities and may miss important visual cues during the encoding process.
In contrast, we use VLM's comprehensive QA task, including perception, prediction, and planning, to ensure that the visual encoder encodes important features in the input signal, such as key objects that need to be avoided, while ignoring some information that is not important for planning, such as some unimportant objects behind the vehicle.

On the other hand, SOLVE-VLM achieves the highest performance compared to other VLM-based methods, surpassing OmniDrive~\cite{wang2024omnidrive} and DriveVLM~\cite{tian2024drivevlm} by 0.05 and 0.03 in terms of L2 error, using large language models of similar capacities (DriveVLM uses QwenVL-7.7B~\cite{bai2023versatile}, OmniDrive and our SOLVE-VLM use LLaVA1.5-7B~\cite{touvron2023llama}). This experiment proves that 1) our proposed SQ-former can effectively compress visual features and better release the reasoning ability of the language model. 2) Our joint training scheme of sharing SQ-Former enables VLM to obtain effective auxiliary supervision from the end-to-end model branch to improve performance.
Finally, using the future trajectory predicted by VLM in the historical frame as the initialization of the current moment, SOLVE-E2E (Async) achieves better performance than SOLVE-E2E, with an improvement of 0.01 in terms of L2 (m) error.

\subsection{Qualitative Results}
We show two planning results of SOLVE in Fig. \ref{fig:vlm_input}. To a better understanding of the scene, we also provide input surrounding camera images and project the planning trajectories to the front camera image. As shown in the figure,  for simple straight-ahead scenarios (see first row), both VLM and the E2E model can predict trajectories that match the expert trajectory. However, for more complex scenarios, the trajectory predicted by the end-to-end model may deviate from the center line of the lane (see yellow line in second row). At the same time, VLM can better understand the scenario and keep the trajectory in the middle of the lane. Besides, with the help of VLM-predicted trajectories, E2E-Async can predict better trajectories than E2E networks.

\subsection{Ablation Studies}
We conduct comprehensive ablation studies to verify the effectiveness of each component in SOLVE. Unless otherwise mentioned, we train both the SQ-Former and VLM for 6 epochs to optimize computational resources while maintaining experimental rigor. We take L2 (cm) as the default metric for comparison.

\noindent
\textbf{Effects of query number of SQ-former.}
Table~\ref{tab:query-number} displays the VLM performances of using different numbers of SQ-former queries to encode visual information. First, too few collector query numbers (\emph{i.e.} 256) can only represent limited features, which may easily lead to the loss of visual features, resulting in the lowest performance. Second, too many collector queries (\emph{i.e.} 512) may introduce redundant information while increasing the capacity of the SQ-former. It will also increase the amount of computation during VLM training and inference. Finally, we choose 384 queries as our default option to balance performance and computation.

\noindent
\textbf{Effects of different designs of visual information encoding in SQ-Former.} 
Table~\ref{tab:different-design} investigates the impact of different visual information encoding strategies for VLM. Q means the initial learnable query and the arrows indicate the order in which SQ-Former encodes different elements, including image features, detection query, and lane query. Comparing the first and second rows of the table, we can find that incorporating the encoding of the overall image features can achieve a 0.7 performance improvement of L2, which proves that some scene-level clues contained in the image features, such as road conditions and weather, are beneficial to VLM's trajectory planning.
Comparing the second and third rows of the table, we find that the encoding order of different elements also affects the performance. Encoding the detection query first and then the lane query can achieve a 0.3 L2 improvement compared to the reverse order. We use the design in the second row as our choice.

\begin{table}[t]
\center
\scalebox{0.9}{
\begin{tabular}{c|cccc}
\toprule
\multirow{2}{*}{Number of SQ-Former's queries} & \multicolumn{4}{c}{L2 (cm) $\downarrow$} \\
                                  & 1s   & 2s  & 3s  & \cellcolor{gray!30}Avg \\ \hline
256                               & 14.1  & 29.2 & 54.5 & \cellcolor{gray!30}32.6 \\ 
384                               & 14.1  & 29.0 & 54.0 & \cellcolor{gray!30}32.3  \\ 
512                               & 14.2  & 29.2 & 54.4 & \cellcolor{gray!30}32.6 \\ 
\bottomrule
\end{tabular}}
\center
\vspace{-5mm}
\caption{Effects of the number of queries of SQ-Former to encode temporal vision information.}
\label{tab:query-number}
\vspace{-1mm}
\end{table}

\begin{table}[!t]
\center
\scalebox{0.9}{
\begin{tabular}{l|cccc}
\toprule
\multirow{2}{*}{Method} & \multicolumn{4}{c}{L2 (cm) $\downarrow$} \\
                                  & 1s   & 2s  & 3s  & \cellcolor{gray!30} Avg \\ \hline
Q → Det →  Lane          & 14.3  & 29.5 & 55.0 & \cellcolor{gray!30} 33.0 \\
Q → Img →  Det →  Lane                  & 14.1  & 29.0 & 54.0 & \cellcolor{gray!30} 32.3 \\ 
Q → Img →  Lane →  Det                  & 14.2  & 29.1 & 54.4 & \cellcolor{gray!30} 32.6 \\  
\bottomrule
\end{tabular}}
\center
\vspace{-3mm}
\caption{Different designs of visual feature encoding in SQ-Former, where Q means the initial learnable query and $\rightarrow$ means the order of cross attention.}
\label{tab:different-design}
\vspace{-3mm}
\end{table}

\noindent
\textbf{Effects of trajectory CoT for VLM planning.}
Table~\ref{tab:cot} investigates the impact of the proposed trajectory CoT for VLM. Results demonstrate that direct trajectory prediction without CoT yields suboptimal performance, highlighting the inherent challenges VLMs face in auto-regressive numerical prediction tasks. 
The integration of trajectory CoT, through reference trajectories and trajectory tokens, provides crucial location priors, reducing the uncertainty of direct waypoint prediction and yielding a 1.1 performance improvement. Our analysis of reference trajectory quantity reveals that insufficient samples (\eg, 4) constrain trajectory diversity, though still achieving a 0.9 improvement over the baseline. Conversely, excessive samples (\eg, 8), while providing more optional initial references, increase both trajectory selection complexity and computational overhead. Based on this trade-off analysis, we empirically determine 6 reference trajectories as the optimal configuration, balancing performance and computational efficiency.

\noindent
\textbf{Effects of sharing SQ-Former.}
Table~\ref{tab:joint-train} shows the impact of shared SQ-Former features on the performance of the E2E planner and VLM planner. The results indicate that feature sharing and joint training protocols significantly outperform independently trained networks. The E2E network exhibits a 1.5 cm reduction in L2 error when utilizing shared SQ-Former features, suggesting that question-answering supervision enables more effective feature extraction for planning compared to conventional detection and segmentation supervision alone. Similarly, VLM performance improves by 0.6 through E2E network integration. This contrasts with DriveVLM's approach of trajectory initialization through post-processing, which shows no significant improvement in VLM's planning capabilities.

\begin{table}[t]
\center
\scalebox{0.9}{
\begin{tabular}{cc|cccc}
\toprule
\multirow{2}{*}{Method} & \multirow{2}{*}{\begin{tabular}[c]{@{}c@{}}Trajectory\\ number\end{tabular}} & \multicolumn{4}{c}{L2 (cm) $\downarrow$} \\
                        &                                                                             & 1s  & 2s  & 3s  & \cellcolor{gray!30}Avg  \\ \hline
\emph{w/o} CoT & - & 13.6   & 26.9   & 47.8   & \cellcolor{gray!30}29.5    \\ \hline
\multirow{3}{*}{CoT}    & 4 (3+1)                                                                          & 12.6   & 25.7   & 47.3   & \cellcolor{gray!30}28.6    \\
                        & 6 (5+1)                                                                          & 12.5   & 25.6   & 47.1   & \cellcolor{gray!30}28.4    \\
                        & 8 (7+1)                                                                          & 12.5   & 25.6   & 47.1   & \cellcolor{gray!30}28.4     \\ 
                        \bottomrule
\end{tabular}
}
\center
\vspace{-5mm}
\caption{Effects of the proposed T-CoT. The trajectory number means the selected top-k and 1 predicted trajectories.}
\label{tab:cot}
\end{table}

\begin{table}[]
\center
\scalebox{0.9}{
\begin{tabular}{cc|cccc}
\toprule
\multirow{2}{*}{Method} & \multirow{2}{*}{SQ-Former} & \multicolumn{4}{c}{L2 (cm) $\downarrow$} \\
                        &                          & 1s     & 2s  & 3s  & \cellcolor{gray!30}Avg  \\ \hline
\multirow{2}{*}{E2E}    & \emph{w/o} sharing              & 14.9   & 29.6 & 53.2 & \cellcolor{gray!30}32.5    \\
                        & sharing                  & 14.2   & 28.1 & 50.5 & \cellcolor{gray!30}31.0    \\ \hline
\multirow{2}{*}{VLM}    & \emph{w/o} sharing              & 12.8   & 26.0   & 48.3   & \cellcolor{gray!30}29.0    \\
                        & sharing                  & 12.6   & 25.5   & 47.3   & \cellcolor{gray!30}28.4    \\ \bottomrule
\end{tabular}}
\center
\vspace{-3mm}
\caption{Effects of sharing SQ-Former features to the E2E module and the VLM module.}
\vspace{-5mm}
\label{tab:joint-train}
\end{table}


\section{Conclusion}
In this work, we introduce SOLVE, an innovative framework that fosters synergy between the Vision-Language Model (VLM) and the end-to-end (E2E) model, emphasizing both knowledge and planning integration. To tackle the challenges faced by VLMs in directly generating fine-grained trajectories in an auto-regressive manner, we present the Trajectory Chain-of-Thought (T-CoT), which progressively refines trajectories using a predefined trajectory bank and chain-style reasoning. Additionally, we propose a temporal decoupling strategy to facilitate cooperation between the VLM and the E2E model. Experiments on the nuScenes dataset demonstrate that our approach achieves state-of-the-art results.

\section{Acknowledgement}
This project is funded in part by National Key R\&D Program of China Project 2022ZD0161100, by the Centre for Perceptual and Interactive Intelligence (CPII) Ltd under the Innovation and Technology Commission (ITC)’s InnoHK, by General Research Fund of Hong Kong RGC Project 14204021, by NSFC-RGC Joint Fund Project N\_CUHK498/24. Hongsheng Li is a PI of CPII under the InnoHK.

{
    \small
    \bibliographystyle{ieeenat_fullname}
    \bibliography{main}
}


\end{document}